\documentclass[runningheads]{llncs}

\usepackage{eccv}

\usepackage{eccvabbrv}

\usepackage{graphicx}
\usepackage{booktabs}

\usepackage[accsupp]{axessibility}  

\usepackage[pagebackref,breaklinks,colorlinks,citecolor=eccvblue]{hyperref}

\usepackage{orcidlink}

\usepackage{multirow} 
\usepackage{pifont}

\begin{document}

\title{UNINEXT-Cutie: The 1st Solution for LSVOS Challenge RVOS Track} 

\titlerunning{The 1st Solution for LSVOS Challenge RVOS Track}

\author{Hao Fang \and
Feiyu Pan \and
Xiankai Lu \and
Wei Zhang \and
Runmin Cong}

\authorrunning{H.~Fang et al.}

\institute{Shandong University\\
Team: MVP-TIME}

\maketitle

\begin{abstract}
Referring video object segmentation (RVOS) relies on natural language expressions to segment target objects in video. In this year, LSVOS Challenge RVOS Track replaced the origin YouTube-RVOS benchmark with MeViS. MeViS focuses on referring the target object in a video through its motion descriptions instead of static attributes, posing a greater challenge to RVOS task. In this work, we integrate strengths of that leading RVOS and VOS models to build up a simple and effective pipeline for RVOS. Firstly, We finetune the state-of-the-art RVOS model to obtain mask sequences that are correlated with language descriptions. Secondly, based on a reliable and high-quality key frames, we leverage VOS model to enhance the quality and temporal consistency of the mask results. Finally, we further improve the performance of the RVOS model using semi-supervised learning. Our solution achieved 62.57 \( \mathcal{J} \)\&\( \mathcal{F} \) on the MeViS test set and ranked 1st place for 6th LSVOS Challenge RVOS Track.
\end{abstract}

\section{Introduction}
\label{sec:intro}
Referring video object segmentation (RVOS) is a continually evolving task that aims to segment target objects in video, referred to by linguistic expressions. 
The 6th LSVOS challenge features two tracks: the VOS track and the RVOS track. This year, the challenge has introduced new datasets, MOSE~\cite{ding2023mose} and MeViS~\cite{ding2023mevis}, replacing the classic YouTube-VOS~\cite{xu2018youtube} and Referring YouTube-VOS benchmarks~\cite{seo2020urvos} used in previous editions. The new datasets present more complex scenes, including scenarios with disappearing and reappearing objects, inconspicuous small objects, heavy occlusions, crowded environments, and long-term videos, making this year's challenge more difficult than ever before.

Most early RVOS approaches~\cite{khoreva2019video,liang2021rethinking} adopt multi-stage and complex pipelines that take the bottom-up or top-down paradigms to segment each frame separately. Meanwhile, compared to rely on complicated pipelines, MTTR~\cite{botach2022end} and Referformer~\cite{wu2022language} first adopt end-to-end framework modeling the task as the a sequence prediction problem, which greatly simplifies the pipeline. Based on the end-to-end architecture of Transformer, SOC~\cite{luo2023soc} and MUTR~\cite{yan2024referred} achieve excellent performance by efficiently aggregating intra and inter-frame information. For example, the 1st and 2nd place solution for MeViS Track in CVPR 2024 PVUW Workshop~\cite{ding2024pvuw,pan20243rd} both involve fine-tuning MUTR~\cite{yan2024referred} on MeViS~\cite{ding2023mevis}. Some solutions also use zero-shot VOS~\cite{lu2020zero,lu2021segment} and VIS~\cite{fang2024learning,fang2024unified} frameworks to handle RVOS task. Meanwhile, UNINEXT~\cite{yan2023universal} proposes a unified prompt-guided formulation for universal instance perception, reuniting previously fragmented instance-level subtasks into a whole and achieve the state-of-the-art performance for the RVOS task.

In this work, we integrate strengths of that leading RVOS and VOS models to build up a simple and effective pipeline for RVOS. Specifically, We finetune RVOS model UNINEXT~\cite{yan2023universal} to obtain mask sequences that are correlated with language descriptions. While UNINEXT can provide masks that are strongly correlated with reference descriptions, but has certain limitations in temporal consistency due to the difficulty of MeViS. Inspired by the previous challenge solutions~\cite{luo20241st,hu20221st}, we leverage the state-of-the-art VOS model Cutie~\cite{cheng2024putting} to enhance the quality and temporal consistency of the mask results based on a reliable and high-quality key frames. Results after Cutie enhancement has performed better than the baseline model, thus the predicted results on the validation set of MeViS dataset can be served as pseudo ground truth object masks of validation set. We then re-finetune the baseline model on validation set with pseudo labels. This semi-supervised approach~\cite{cao2022second} is also employed on the testing set.

In this year, the 6th LSVOS Challenge RVOS Track replaced the origin YouTube-RVOS benchmark with MeViS~\cite{ding2023mevis}. Motion Expression guided Video Segmentation requires segmenting objects in video content based on sentences describing object motion, which is more challenging compared to traditional RVOS datasets. Our solution achieved 58.93 \( \mathcal{J} \)\&\( \mathcal{F} \) on the MeViS val set, 62.57 \( \mathcal{J} \)\&\( \mathcal{F} \) on the MeViS test set and ranked 1st place for 6th LSVOS Challenge RVOS Track.

\begin{figure}[t]
\begin{center}
\includegraphics[width=1\linewidth]{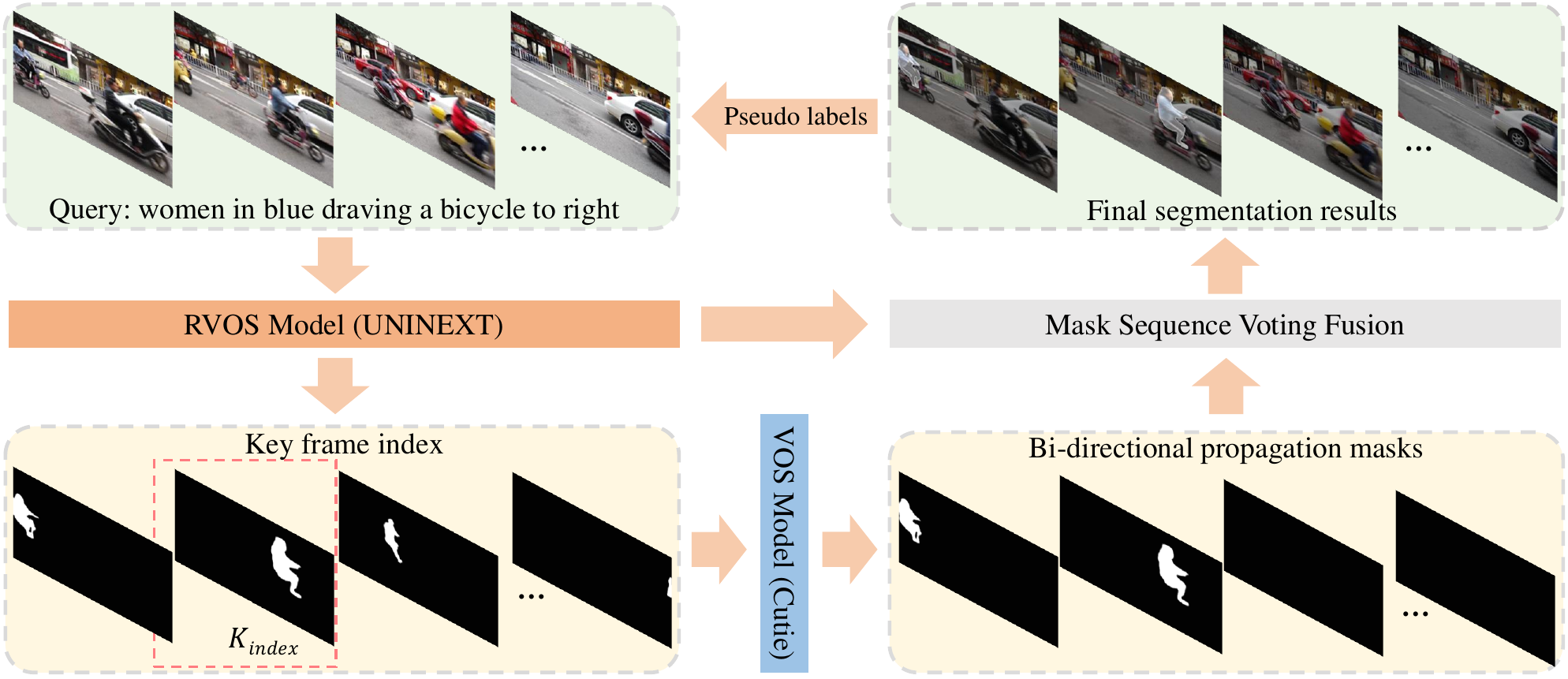}
\end{center}
\caption{The overview architecture of the proposed method.}
\label{fig:model}
\end{figure}

\section{Method}
\label{sec:metho}
The input of RVOS contains a video sequence $\mathcal{V} = \left\{v_t\in \mathbb{R}^{3 \times H \times W} \right\}_{t=1}^T $ with \textit{T} frames and a corresponding referring expression $\mathcal{E} = \left\{ e_l \right\}_{l=1}^L $ with \textit{L} words. Our solution consists of three steps: Backbone(~\cref{sec:backbone}), Post-process(~\cref{sec:post-process}), and Semi-supervised(~\cref{sec:semi-supervised}). The overall architecture of the proposed method is illustrated in ~\cref{fig:model}. 

\subsection{Backbone}
\label{sec:backbone}
We adopt the state-of-the-art RVOS model UNINEXT~\cite{yan2023universal} as our backbone to obtain mask sequences $\mathcal{S} = \{s_t\}_{t=1}^T$ that are correlated with language descriptions.
\begin{equation}
    \mathcal{S} = \mathcal{F}^{rvos}\left( \mathcal{V}, \mathcal{E}\right),
\end{equation}
where $\mathcal{F}^{rvos}$ denotes the UNINEXT model. UNINEXT reformulates diverse instance perception tasks into a unified object discovery and retrieval paradigm, and achieved surprising performance after joint training on multiple datasets. So we fine-tuned the official pre-training weights provided on MeViS.

\subsection{Post-process}
\label{sec:post-process}
Previous challenge solutions~\cite{luo20241st,hu20221st} have shown that using a semi-supervised VOS algorithm can further improve the accuracy of segmentation results. The general procedure are first selecting the key-frame index of mask sequences probability $\mathcal{P}$ from RVOS model, then using VOS model to perform forward and backward propagation. It can be formulated as:
\begin{equation}
\begin{gathered}
    \mathcal{K}_{index} = argmax(\mathcal{P}), \\
    \mathcal{M} = \left[\mathcal{F}^{vos}\left(\{s_{i}\}_{i=K_{index}}^{0}\right), \mathcal{F}^{vos}\left(\{s_{j}\}_{j=K_{index}}^{T}\right)\right], \\
\end{gathered}
\end{equation}
where $\mathcal{P} = \left\{ p_k \in \mathbb{R}^1 \right\}_{k=1}^T$, $\mathcal{F}^{vos}$ denotes the VOS model for post-process. We adopt the state-of-the-art VOS model Cutie~\cite{cheng2024putting} for post-process.

In our experiment, we find that post-process dose not improve the mask quality of all videos. The reason is that MeViS is a multi-object dataset, and the mask with the highest probability output by UNINEXT may not necessarily include all specified objects. This may not be a problem with UNINEXT, it could just be that only a single object appeared in that frame. Therefore, we select the $N$ masks with the highest probability in the RVOS model for VOS inference and fuse them with the mask sequence output by the original RVOS model.
\begin{equation}
    \mathcal{M} = \mathcal{F}^{fuse}\left( \mathcal{S}, \mathcal{M}^{N}\right),
\end{equation}
where $\mathcal{M}^{N}$ is the $N$ sets of mask sequences output by Cutie, $\mathcal{F}^{fuse}$ denotes pixel-level binary mask voting. If there are more than $(N+1)/2$ pixels with a value equal to 1, we divide the pixel into the foreground, otherwise, it is divided into the background.

\subsection{Semi-supervised}
\label{sec:semi-supervised}
The post-processing result $\mathcal{M}$ is significantly better than the backbone result $\mathcal{S}$, thus the predicted results on the validation set of MeViS dataset can be served as pseudo ground truth object masks of validation set. We then re-finetune the backbone model UNINEXT on validation set with pseudo labels. This semi-supervised approach~\cite{cao2022second} is also employed on the testing set. Finally, performing further post-processing after fine-tuning can further improve performance.

\begin{figure}[t]
\begin{center}
\includegraphics[width=1\linewidth]{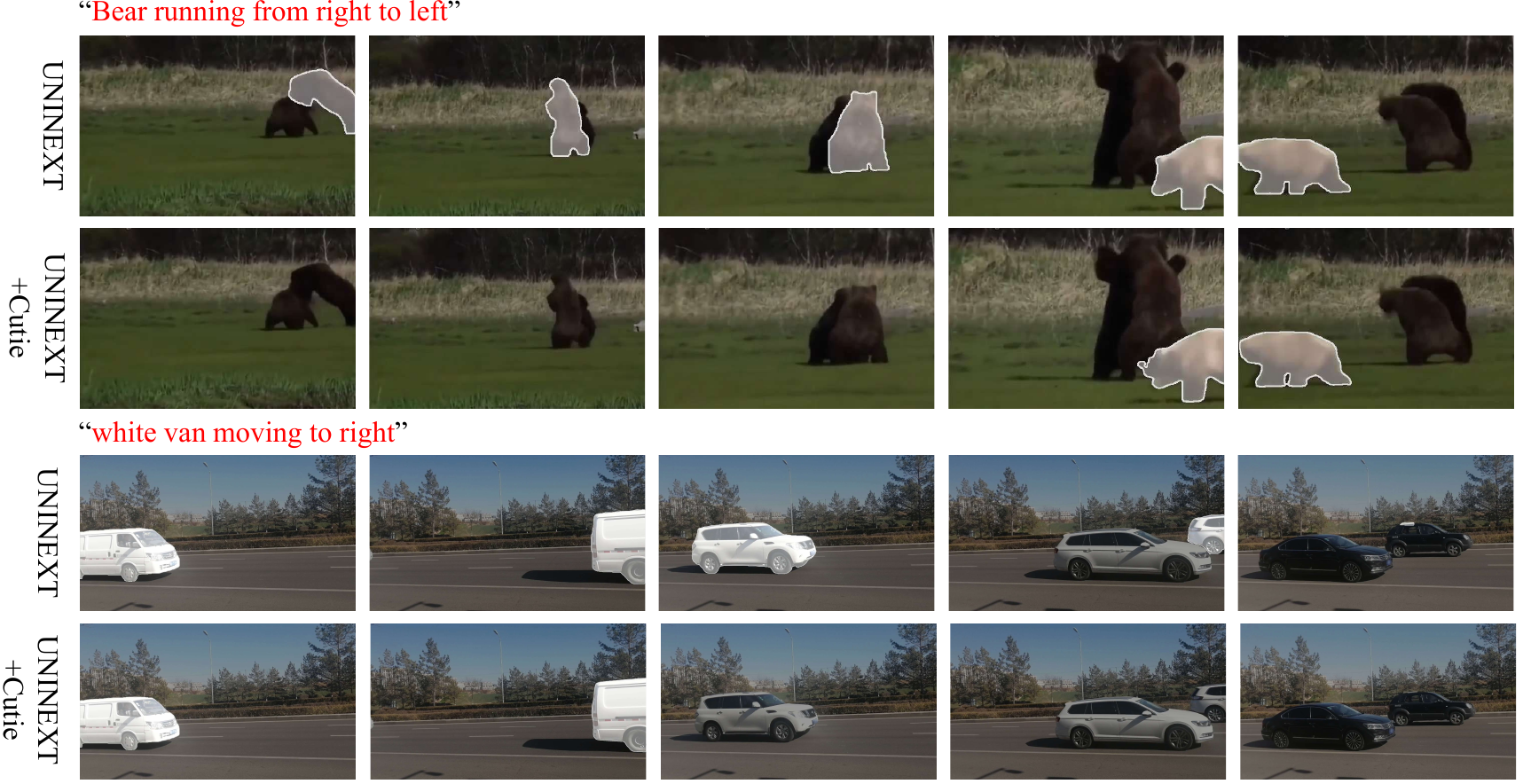}
\end{center}
\caption{Qualitative comparison on validation dataset.}
\label{fig:rvos}
\end{figure}

\section{Experiment}
\label{sec:exper}

\subsection{Datasets and Metrics}
\noindent \textbf{Datasets.} 
MeViS~\cite{ding2023mevis} is a newly established dataset that is targeted at motion information analysis and contains 2,006 video clips and 443k high-quality object segmentation masks, with 28,570 sentences indicating 8,171 objects in complex environments. All videos are divided into 1,662 training videos, 190 validation videos and 154 test videos.

\noindent \textbf{Evaluation Metrics.} 
we employ region similarity $\mathcal{J}$ (average IoU), contour accuracy $\mathcal{F}$ (mean boundary similarity), and their average \( \mathcal{J} \)\&\( \mathcal{F} \) as our evaluation metrics.

\subsection{Implementation Details}
We adopt the pre-trained weights of UNINEXT as initialization and fine-tune model on MeViS. We use ViT-Huge~\cite{dosovitskiy2020image} as the visual encoder and BERT~\cite{devlin2018bert} as the text encoder, both of which are frozen during fine-tuning. We use the AdamW optimizer with a learning rate of 1e-4, a batch size of 8, and a weight decay of 0.05. The training lasts for 50K iterations, with the learning rate reduced by 10 times after 40K iterations. In post-process, we directly adopt Cutie-base+ weights trained on MEGA~\cite{cheng2024putting} for zero-shot inference. In semi-supervised learning, we re-finetune UNINEXT on training-set and joint with validation-set with pseudo ground truth. The experimental setup is consistent with the first fine-tuning.

\subsection{Main Results}
As shown ~\cref{tab:leader}, our approach achieves 62.57\% in \( \mathcal{J} \)\&\( \mathcal{F} \), ranking in 1st place for 6th LSVOS Challenge RVOS Track.

\begin{table}
\caption{The leaderboard of the MeViS test set.}
\setlength\tabcolsep{10pt}
  \centering
  \begin{tabular}{l|ccc}
    \toprule
    Team & \( \mathcal{J} \)\&\( \mathcal{F} \) & $\mathcal{J}$ & $\mathcal{F}$ \\
    \midrule
    \bf MVP-TIME & \bf 62.57 & \bf 58.98 & \bf 66.15\\
    TXT & 60.40 & 57.02 & 63.78\\
    CASIA\_IVA & 60.36 & 56.88 & 63.85\\
  \bottomrule
\end{tabular}
\label{tab:leader}
\end{table}

\subsection{Ablation Study}
To validate the effectiveness of each step, we conduct simple ablation studies. As shown in ~\cref{tab:ablation}, The fine-tuning of UNINEXT has already achieved 50.51\% in \( \mathcal{J} \)\&\( \mathcal{F} \). Cutie's post-process brings 4.83\% in \( \mathcal{J} \)\&\( \mathcal{F} \) of improvement. The semi-supervised approach further improves by 3.59 \% in \( \mathcal{J} \)\&\( \mathcal{F} \).
\begin{table}
\setlength\tabcolsep{10pt}
\caption{Ablation study on the MeViS validation set.}
  \centering
  \begin{tabular}{c|ccc}
    \toprule
    Method & \( \mathcal{J} \)\&\( \mathcal{F} \) & $\mathcal{J}$ & $\mathcal{F}$ \\
    \midrule
    UNINEXT & 50.51 & 47.28 & 53.74\\
    + Cutie & 55.34\textcolor{red}{(+4.83)} & 52.49 & 58.19\\
    + Semi-supervised & 58.93\textcolor{red}{(+3.59)} & 56.39 & 61.46\\
  \bottomrule
\end{tabular}
\label{tab:ablation}
\end{table}

\subsection{Qualitative Results}
~\cref{fig:rvos} shows some visualization results, where Cutie further enhance the quality and temporal consistency of the mask results.

\section{Conclusion}
In this work, we integrate strengths of that leading RVOS and VOS models to build up a simple and effective pipeline for RVOS. We finetune RVOS model to obtain mask sequences that are correlated with language descriptions. Based on a reliable and high-quality key frames, we leverage VOS model to enhance the quality and temporal consistency of the mask results. We further improve the performance of the RVOS model using semi-supervised learning. Our solution achieved 62.57 \( \mathcal{J} \)\&\( \mathcal{F} \) on the MeViS test set and ranked 1st place for 6th LSVOS Challenge RVOS Track.

\bibliographystyle{splncs04}
\bibliography{main}

\end{document}